\pdfoutput=1

\documentclass[11pt]{article}

\usepackage{EMNLP2023}

\usepackage{times}
\usepackage{latexsym}

\usepackage[T1]{fontenc}

\usepackage[utf8]{inputenc}

\usepackage{microtype}

\usepackage{inconsolata}

\usepackage{amsmath}
\usepackage{algorithm}
\usepackage{algorithmic}
\usepackage{graphicx}
\usepackage{booktabs} 

\title{Cross-Lingual Supervision improves Large Language Models Pre-training}

\author{Andrea Schioppa \\
  Google Research \\
  \texttt{arischioppa@google.com} \\\And
  Xavier Garcia \\
  Google Research \\
  \texttt{xgarcia@google.com} \\\And 
  Orhan Firat \\
  Google Research \\
  \texttt{orhanf@google.com} 
  \\}

\begin{document}
\maketitle
\begin{abstract}
The recent rapid progress in pre-training Large Language Models has relied on using self-supervised language modeling objectives
like next token prediction or span corruption. On the other hand, Machine Translation Systems are mostly trained using
cross-lingual supervision that requires aligned data between source and target languages. We demonstrate that \emph{pre-training}
Large Language Models on a mixture of a self-supervised Language Modeling objective and the supervised Machine Translation objective, therefore \emph{including cross-lingual parallel data during pre-training}, yields models with better
in-context learning abilities. As pre-training is a very resource-intensive process and a grid search on the
best mixing ratio between the two objectives \emph{is prohibitively expensive}, we propose a simple yet effective strategy to  learn it during pre-training.
\end{abstract}
\section{Introduction}
The rapid progress in the development of large-scale pre-training, GPT~\cite{gpt3}, XGLM~\cite{xglm}, PaLM~\cite{palm}, has resulted in models capable of performing a variety of tasks through the \emph{in-context learning (aka.~few shot) paradigm}~\cite{gpt3}: one can present the model a few demonstrations of a given task at inference, and the model will able to follow these demonstrations on \emph{new, unseen examples}. Therefore it is \emph{no longer necessary to fine-tune} these models on a variety of down-stream tasks. The pre-training of such Large Language Models (LLMs) \emph{relies on self-supervision}, i.e.~the data does not require to be annotated. Examples of self-supervised (LM) Language Modeling objectives are next token prediction, where the task is to predict the next token given the previous ones, or span-corruption where the task is to fill-in a portion of missing text given its surroundings.

On the other hand, Machine Translation Models (MTMs) are still being trained using cross-lingual supervision, which \emph{requires aligned parallel data}. Indeed, the Machine Translation (MT) objective consists in predicting the target sentence
given the source sentence, and therefore it is \emph{necessary to collect aligned pairs} of texts between source and target
languages.

On Machine Translation, pre-trained LLMs have historically under-performed MTMs trained just on millions of supervised examples both when the LLMs are \emph{evaluated using in-context learning}, or \emph{after having been fine-tuned on parallel data}. However, the performance gap between LLMs and MTMs has been shrinking. For example, the recent PaLM~\cite{palm}, a language model pre-trained using self-supervision only, is able to outperform previous state-of-the-art MTMs on older machine translation benchmarks,
while still lagging behind supervised MTMs on recent benchmarks~\cite{palm_mt_prompt}. Such a trend raises the natural question Q: \emph{Is training on cross-lingual supervised data still necessary or beneficial?}

Regarding question Q, we think that the most promising direction to explore is the \emph{inclusion of parallel data when pre-training LLMs}.
The first rationale for our preference is the shrinking gap on MT benchmarks between LLMs and MTMs: it is quite likely
that LLMs will be able to catch-up in the nearby future, while at the same time being able to perform many more tasks than MTMs. The second rationale is that pre-training datasets are still dominated by English, compare the language composition of the pre-training dataset for PaLM~\cite{palm}: other languages, especially lower resource ones, \emph{are under-represented}.
Therefore, a natural conjecture is that aligned cross-lingual data \emph{might enhance the abilities of LLMs across languages other than English}. 

When assessing the multi-lingual abilities of LLMs we need to distinguish between the \emph{open and closed generation}
settings. In closed generation the task \emph{is performed in a single language}; for example a context paragraph is
presented in German, questions are formulated in German and answers are expected in German. In open generation the
task \emph{is performed across two languages}; for example a context paragraph is presented in English, questions are formulated in German and answers are expected in German. Now the attractiveness of including cross-lingual data during pre-training lies not only in the ability to improve the machine translation performance of LLMs, but also in \emph{building a bridge between
languages}. While we might expect that cross-lingual supervision improves closed generation in under-represented languages,
another natural conjecture is that it improves open generation, i.e.~where two languages are involved. 

In light of the above discussion we refine Q: \emph{Is cross-lingual supervised data beneficial when pre-training LLMs? In particular, are there gains both on open and closed generation when using the in-context learning paradigm for evaluation?}
 
We are not the first to consider the usage of cross-lingual supervision with LLMs, see Section~\ref{sec:related_work}. However our study differs from previous ones in the following aspects:
\begin{enumerate}
    \item We include cross-lingual supervision at the \emph{pre-training stage}.
    \item We include cross-lingual supervision using the standard supervised MT objective.
    \item We evaluate the resulting models with in-context learning considering \emph{both closed and open generation settings}.
    \item We \emph{learn} the amount of parallel data to use \emph{while training}.
\end{enumerate}

In this work we first demonstrate that including \emph{some cross-lingual supervision is beneficial when pre-training large language models}, thus answering Q. Then, when faced with learning an optimal amount of cross-lingual supervision to use, we show that automated curriculum learning~\cite{graves_curriculum} is \emph{an effective strategy that does not require multiple training runs and which outperforms static policies}.

We emphasize the importance of learning the amount of parallel data while training \emph{without resorting to a hyper-parameter search}. Pre-training an LLM on sufficiently many tokens is a \emph{resource intensive task}; for example each of our experiments with a 3.8B-parameter models requires 256 TPUv4 cores for 5 days. If we treat the mixing ratio between the parallel MT data and the LM training data as a hyper-parameter $\lambda$, we have \emph{in theory just an additional hyper-parameter}. However a grid search is \emph{prohibitively expensive}; for example~\cite{kale-etal-2021-nmt5} considered a \emph{less compute-intensive}
setup in which one \emph{fine-tunes mT5 models for 100k steps} on a mixture of MT and LM data; nevertheless they were able to
just compare \emph{two values of $\lambda$}. Furthermore, treating $\lambda$ as a hyper-parameter overlooks the fact that
there might be \emph{dynamic scheduling strategies}, i.e.~varying $\lambda$ over time, that outperform static ones in which $\lambda$ is held fixed.

\section{Related work}\label{sec:related_work}
We are not the first to investigate the usage of parallel cross-lingual data with LLMs.
\cite{reid-artetxe-2022-paradise} considered leveraging parallel data by devising a loss consisting of 3 objectives; however, their technique is somewhat complicated because it necessitates the development of a multi-lingual noising procedure, while
we opt for including the cross-lingual data using the standard MT objective.
\cite{chi-etal-2021-mt6} proposed a simpler objective by building on top of the success of \cite{mt5}: directly adding supervised MT data to the denoising procedure used for training mT5, which results in models outperforming mT5 in cross-lingual generation.
Note however, that while \cite{chi-etal-2021-mt6} includes cross-lingual supervision during pre-training, the
resulting models \emph{do not display in-context learning abilities} and evaluation is carried out by fine-tuning on
down-stream tasks. \cite{kale-etal-2021-nmt5} explored what happens by fine-tuning mT5 on parallel data; therefore
parallel data is used during an \emph{intermediate stage between pre-training and fine-tuning on down-stream tasks}.
A limitation of all these studies is the emphasis on \emph{fine-tuning}: all of these models require fine-tuning, which is quite different from few-shot in-context learning. As such, the question of whether supervised data in one task can benefit few-shot learning in another task \emph{remains unexplored}.

\section{Basic Setup}
\subsection{Training Data}
Our Language Modeling data is based on that from \cite{palm} but we slightly modify the proportions between different sub-categories, see Table~\ref{tab:lm_data}. We do not use a public Language Modeling dataset, e.g.~MC4~\cite{t5paper}, as in early experiments the high-quality data from~\cite{palm} yielded  better in-context learning abilities. As Language Modeling objective we use the recent ``UL2''~\cite{https://doi.org/10.48550/arxiv.2205.05131} because it has shown better performance in the few-shot setting.

\par For the MT data we use an in-house parallel corpus covering the languages in Table~\ref{tab:mt_data}, which also reports the
sampling proportions and highlights whether we consider a language in the High or Low resource setting. 
Note that our training data has always the source or the target in English. We use the
standard approach used when training multi-lingual supervised models:
\begin{align}\label{eq:mt_ingestion}
    \langle 2xx \rangle + \rm{source} \to\ & \rm{encoder} \\
    \rm{target} \to\ & \rm{decoder},
\end{align}
where the source sentence is prefixed with a special target language token, $\langle 2xx \rangle$, and is supplied to the Encoder, while the target is supplied to the Decoder.

\begin{table}[t]
\centering
\begin{small}
\begin{tabular}{|l|r||}
\toprule
Data Source. & \% of Data \\
\midrule
Social media $\text{conversations}^{\dag}$ & 40\% \\
Filtered $\text{webpages}^{\dag}$ & 34\% \\
GitHub & 4\% \\
$\text{Books}^{*}$ & 15\% \\
$\text{Wikipedia}^{\dag}$ & 5\% \\
$\text{News}^*$ & 2\% \\
\bottomrule
\end{tabular}
\end{small}
\caption{LM Data: Data sources and proportion of data. $\dag$ means the data source is multilingual, while $*$ means it is English-only.}
\label{tab:lm_data}
\end{table}
\begin{table}[t]
\centering
\begin{small}
\begin{tabular}{|l|r|r|r|}
\toprule
Data Source. & \% of Data & Low/High & Tokens (B) \\
\midrule
Sentences & 96\%  & - & -\\
Documents & 4\% & - & - \\
\midrule
ar & 7.3\%  & High & 16.5 \\
bn & 5.4\% & Low & 1.4 \\
de & 9.5\% & High & 85.7 \\
fi & 6.3\% & High & 8.2 \\
fr & 9.8\% & High & 123.6 \\
id & 7.1\% & High & 5.8 \\
ja & 7.9\% & High & 27.9 \\
ko & 7.2\% & High & 15.8 \\
ru & 8.6\% & High & 54.6 \\
sw & 4.8\% & Low & 1.1 \\
te & 4.5\% & Low & 0.5 \\
th & 6.6\% & High & 14.3 \\
tr & 7.9\% & High & 11.9 \\
vi & 7.1\% & High & 12.7 \\
\bottomrule
\end{tabular}
\end{small}
\caption{MT Data composition}
\label{tab:mt_data}
\end{table}

\subsection{Model architecture}
Commonly used LLM architectures are Encoder-Decoder models, e.g.~T5~\cite{t5paper}, and Decoder-only models, e.g.~\cite{gpt3,palm}. Most supervised MTMs use an Encoder-Decoder architecture. As our experiments
require \emph{pre-training from scratch} and are therefore quite resource-demanding, we \emph{consider only one
architecture, the Encoder-Decoder}. Specifically, we use the mT5~\cite{mt5} architecture at model sizes ``large'' (1.2 billion) and ``xl'' (3.8 billion). We train the
1.2B models for 250k steps and the 3.8B billion models for 500k steps using the default settings from the T5X library \cite{roberts2022t5x}. In our
batches the maximum sequence length is 1024 and the number of non-padding tokens is slightly over 500k. We emphasize that
\emph{mT5 is only used for the architecture, and we never use mT5 checkpoints or the data used to train mT5}.

\subsection{Evaluation} \emph{We evaluate our models with in-context learning using the one-shot setting, see the Appendix for
explicit examples}; concretely, each test input is prefixed with one example displaying the desired input to target behavior;
the sequence thus obtained is supplied to the Encoder and the target is generated by the Decoder.

We consider three tasks: Question Answering, Machine Translation and Summarization. For Question Answering we consider two settings: \emph{closed generation}, where the context, the question
and the answer are in the same language, and \emph{open generation} where the context is in one language and the question and the answer are in another one.  For the closed generation setting we use TyDiQA~\cite{clark-etal-2020-tydi}. For the open generation setting we take the non-English splits of TyDiQA and translate the context to English using the Google Translate API (\url{translate.google.com} accessed in November 2022.); we denote the dataset thus obtained as XTyDiQA. 
For Machine Translation we use Flores~\cite{guzman-etal-2019-flores} and for Summarization we employ Wikilingua~\cite{ladhak-etal-2020-wikilingua}, with the splits and pre-processing from the GEM~\cite{gem_benchmark} benchmark.

\section{Learning to schedule the two tasks}
\textbf{A grid search on $\lambda$ is unfeasible.} As we have two tasks, Language Modeling and Machine Translation,
we might treat the proportion $\lambda$ of the MT task as a hyper-parameter to tune. Given that pre-training is
very resource intensive, a \emph{grid search on $\lambda$ is unfeasible}. Even in the less compute intensive setting
considered by~\cite{kale-etal-2021-nmt5}, which is a continued pre-training of mT5 checkpoints, they were able to
compare \emph{just two values of $\lambda$}. It is therefore highly desirable to learn $\lambda$ during training, with
the additional benefit that a policy changing $\lambda$ over time might outperform one that holds it constant.

\textbf{Automated curriculum learning is a natural approach.} 
When training a model on data from multiple sources, the \emph{automated curriculum learning paradigm}~\cite{graves_curriculum} can learn the data-sampling schedule while training. In this way we can learn a dynamic lambda, $\lambda_t$, which is
a function of the time step $t$; concretely, $\lambda_t$ represents the probability of sampling the MT task and $1-\lambda_t$
is the probability of sampling the LM task.
Recent work~\cite{kreutzer-etal-2021-bandits-dont} has shown promising results when applying this curriculum approach to
Machine Translation Systems where the data comes from multiple domains or multiple languages. For example, \cite{kreutzer-etal-2021-bandits-dont} demonstrates that the multi-armed bandits employed by automated curriculum learning
perform competitively against several SOTA heuristics on multi-lingual benchmarks. 

\textbf{We need to find the right reward function.} In order to learn the dynamic scheduling of the MT and LM tasks, we need
to assign a reward for using a specific task. Suppose that we sample a task $\tau\in\{\rm{MT},\rm{LM}\}$; we then
obtain a corresponding batch $B_\tau$ and perform gradient descent updating the model parameters from $\Theta$
to $\Theta'$. The specific choice of $\tau$ has therefore resulted in a parameter change, \emph{and we need to measure
how useful it was}. After bench-marking different utility functions, \citet{kreutzer-etal-2021-bandits-dont} recommends to measure the loss reduction $L(\Theta) - L(\Theta')$ on a \emph{trusted validation set}. However, while in the setup of \citet{kreutzer-etal-2021-bandits-dont} there is a clear choice of the validation set, we are interested in pre-training
of an LLM that is then applied to down-stream tasks using the in-context learning paradigm. Therefore, it is not trivial to build a validation set representative of all the possible few-shot tasks. In particular, mitigation strategies would be needed to avoid over-fitting to a specific selection of tasks.

\textbf{We use an intrinsic reward function.} In early experiments we contrasted the rewards assigned by each downstream
task (e.g.~Question Answering) with those assigned by the training tasks and found that the signal from the former was smaller in magnitude and had a bigger variance. \emph{We therefore propose to measure rewards intrinsically on the (pre)-training data itself}. Formally, after taking a gradient step on $B_\tau$, we sample with equal probability a reward task $\rho\in\{\rm{MT},\rm{LM}\}$ and obtain a new batch $B_\rho$ on which we measure the loss reduction. We assign equal
probability to each reward task as we do not want to fix a preference of one task over the other. \emph{One clear benefit of
using an intrinsic reward function is that it is no longer necessary to construct a validation dataset}. 
While the usage of the training tasks themselves has been considered in~\cite{graves_curriculum, kreutzer-etal-2021-bandits-dont},
they measure rewards on the \emph{same batch $B_\tau$ used for taking a gradient step}, while we sample an independent batch $B_\rho$, possibly from another task. As $\rho$ is sampled with $50\%$ probability to be equal to $\tau$ and with $50\%$ probability to be equal to the other task, we \emph{measure both task-specific learning and cross-task transfer}.

\begin{table*}
\centering
\begin{small}
\begin{tabular}{|l|c|c|c|c|c|}
\toprule
Model size (B) & Data Selection & TyDiQA En & TyDiQA Non-En & TyDiQA & XTyDiQA \\
\midrule
1.2 & LM (100\%) & 40.23 & 23.76 & 25.59 & 10.40 \\
1.2 & LM (90\%) -- MT (10\%) & 39.77 & 25.03 & 26.67 & 11.07 \\
1.2 & LM (50\%) -- MT (50\%) & 41.59 & 29.42 & 30.78 & 13.75 \\
1.2 & WARMUP & 39.31 & 23.66 & 25.40 & 12.71 \\
1.2 & EXP3 & \textbf{42.50} & 30.00 & 31.39 & 16.54 \\
1.2 & FAIR & 41.14 & \textbf{31.08} & \textbf{32.19} & \textbf{18.85} \\
\midrule
3.8 & LM (100\%) & 47.72 & 32.97 & 34.61 & 13.96 \\
3.8 & EXP3 & \textbf{50.23} & \textbf{42.54} & \textbf{43.39} & 25.82 \\
3.8 & FAIR & 47.50 & 36.65 & 37.85 & \textbf{26.16} \\
\bottomrule
\end{tabular}
\end{small}
\caption{Performance on TyDiQA and XTyDiQA, measured with EM. Static data selection
strategies are outperformed by our automated curriculum. Adding parallel data does not hurt performance on En and significantly improves closed generation performance on other languages and (cross-lingual) open generation.}
\label{tab:qa_results}
\end{table*}

\textbf{The loss reduction needs to be rescaled.} Note that the loss scales for LM and MT can be different during training,
and so the absolute loss decrease $L(\Theta) - L(\Theta')$ is affected by the task used to compute $L$. Indeed, in Machine Translation all information content is given in the source sequence and therefore the perplexity of a translation task is generally lower than that of a language modeling task. We solve this problem by computing the reward as \emph{the relative loss reduction}
\begin{equation}\label{eq:pgnorm_def}
    \rm{reward} = 1 - \frac{L(\Theta', B_{\rho})}{L(\Theta, B_{\rho})},
\end{equation}
which was called ``pgnorm'' in \cite{kreutzer-etal-2021-bandits-dont}.

\textbf{Classical bandit algorithms tend to sample from a single task.}  The policy from sampling from the two tasks is then learned using multi-armed bandits~\cite{lattimore_szepesvaari_2020}. We initially experimented with EXP3 as in \cite{graves_curriculum, kreutzer-etal-2021-bandits-dont}. We discovered, however, that the LM task always produces slightly greater reward than the MT task. As EXP3 is designed to pick the best single arm in hindsight, it tends to center the policy on the LM arm. To mitigate this issue,
we propose a ``FAIR'' algorithm that samples proportionally to a moving average of the rewards for a given arm,
see Algorithm~\ref{alg:fair} for details.
\begin{algorithm}[t]
\caption{FAIR}\label{alg:fair}
\begin{algorithmic}[1]
\REQUIRE exploration rate $\gamma$, moving average rate $\mu$, number of arms $n$
\STATE Initialize arm weights: $w_a\gets 10^{-7}$
\STATE Compute policy: $\pi_a = (1-\gamma) \frac{w_a}{\sum_a w_a} + \frac{\gamma}{n}$
\STATE Sample arm: $a\sim\pi$ and get reward $r_a$
\STATE Update weights: $w_a\gets (1-\mu) w_a + \mu r_a$
\end{algorithmic}
\end{algorithm}
For reproducibility, we provide full details on our curriculum setup in the Appendix.

\section{Experimental results}
\subsection{Baselines}
The first baseline we consider is training on just the LM data (LM (100\%)); as a grid search on a static mixing ratio $\lambda$ between the LM and MT tasks is prohibitively expensive, we consider the two values of $\lambda$ from~\cite{kale-etal-2021-nmt5}:
$\lambda = 0.5$ (LM (50\%) -- MT (50\%)) and $\lambda = 0.1$ (LM (90\%) -- MT (10\%)). To create an intermediate behavior
between $\lambda = 0.5$, which samples the MT objective more aggressively, and $\lambda = 0.1$, which samples it more conservatively, we consider a WARMUP heuristic that uses $\lambda = 0.4$ for the first 20k steps and then defaults to $\lambda = 0.1$. The value $\lambda = 0.4$ was chosen by inspecting the rewards of each task at the beginning of training. 

At model size 1.2B we found that adding parallel data improves performance across the evaluated tasks; however, the automated curriculum learning strategies outperform the other baselines; \emph{thus, given the limited experimental budget, at model size 3.8B we just consider the LM (100\%) as a baseline}.
\subsection{Question Answering}
Our results for Question Answering are in Table~\ref{tab:qa_results}. \emph{For TyDiQA we see that adding parallel data can significantly improve performance on the non-English part and does not degrade the performance on English}. On XTyDiQA, we observe that adding parallel data can make a significant difference with up to +8 EM points at model size 1.2 billion and +12 EM points at 3.8 billion parameters. Therefore, we see that \emph{including cross-lingual supervision during pre-training improves the open generation abilities of the resulting pre-trained models} for the Question Answering task. We also see that our automated curriculum (either EXP3 or FAIR) outperforms all the other data-sampling strategies at model size 1.2B, and therefore we just experiment with EXP3 and FAIR at the larger model size 3.8B.

\begin{table*}
\centering
\begin{small}
\begin{tabular}{|l|c|c|c|c|c|}
\toprule
Model size (B) & Data Selection & En & Non-En & All \\
\midrule
1.2 & LM (100\%) & \textbf{16.11} & 12.37 & 12.71 \\
1.2 & LM (90\%) -- MT (10\%) & 15.99 & \textbf{12.64} & \textbf{12.94} \\
1.2 & LM (50\%) -- MT (50\%) & 14.92 & 11.76 & 12.04 \\
1.2 & WARMUP & 15.21 & 11.82 & 12.13 \\
1.2 & EXP3 & 15.80 & 12.25 & 12.57 \\
1.2 & FAIR & 14.45 & 11.55 & 11.82 \\
\midrule
3.8 & LM (100\%) & 16.2 & 12.32 & 12.67 \\
3.8 & EXP3 & 17.08 & 13.38 & 13.72 \\
3.8 & FAIR & \textbf{18.15} & \textbf{14.15} & \textbf{14.51} \\
\bottomrule
\end{tabular}
\end{small}
\caption{Summarization performance evaluated with RougeL on Wikilingua. At 1.2B parameters, adding more parallel data can slightly decrease performance. However, at 3.8B parameters, adding parallel data slightly improves over the LM-only baseline.}
\label{tab:summarization_results}
\end{table*}

\subsection{Summarization}

Table~\ref{tab:summarization_results} shows the key outcomes for the summarization task on Wikilingua. We contrasted automated curriculum-based techniques to manual mixing methods. Compared to Question Answering, summarization results sway, albeit not significantly, towards larger models where automated curriculum methods outperform the vanilla LM only method and our proposed FAIR method outperforms the others. Interestingly, we did not observe any gains when scaling from 1.2 billion to 3.8 billion parameters for LM (100\%), where the automated curriculum methods benefits from scaling model size more than the than vanilla LM only method.

\begin{table*}
\centering
\begin{small}
\begin{tabular}{|l|c|c|c|c|c|}
\toprule
Model size (B) & Data Selection & En $\rightarrow$ High & High $\rightarrow$ En & En $\rightarrow$ Low & Low $\rightarrow$ En \\
\midrule
1.2 & LM (100\%) & 8.96 & 15.76 & 0.65 & 3.01 \\
1.2 & LM (90\%) -- MT (10\%) & 12.00 & 20.80 & 2.00 & 5.99 \\ 
1.2 & LM (50\%) -- MT (50\%) & 17.74 & 27.14 & 5.22 & 13.94 \\
1.2 & WARMUP & 10.71 & 21.80 & 1.28 & 6.46 \\
1.2 & EXP3 & 16.05 & 26.20 & 4.63 & 18.07 \\
1.2 & FAIR & \textbf{23.19} & \textbf{31.81} & \textbf{15.38} & \textbf{26.73} \\
\midrule
3.8 & LM (100\%) & 12.43 & 21.02 & 1.08 & 5.08 \\
3.8 & EXP3 & 26.63 & 34.88 & 23.63 & 31.53 \\
3.8 & FAIR & \textbf{30.48} & \textbf{36.63} & \textbf{27.53} & \textbf{36.05} \\
\bottomrule
\end{tabular}
\end{small}
\caption{Performance on MT tasks (Flores) measured with sacreBLEU. Adding parallel data greatly improves translation results (evaluated with in-context learning), with the FAIR bandit being the best data selection strategy.}
\label{tab:mt_results}
\vskip -0.1in
\end{table*}

\subsection{Machine Translation}
\textbf{Curriculum learning boosts performance.} For Machine Translation (Table~\ref{tab:mt_results}), we partition our analysis into four settings, into- and out-of-English translation (X$\rightarrow$En, En$\rightarrow$X), and high and low resource translation. \emph{We observe that the gains of using an automated curriculum method can be quite substantial}, with significant gains, e.g.~+10 BLEU points, over the LM (50\%) -- MT (50\%) sampling in the En $\rightarrow$ Low setting. Compared to other methods, our proposed FAIR algorithm also boosts the generation quality further. Note that, given our constrained experimental budget we thus only considered automated curriculum strategies at model size 3.8B.

\begin{table*}
\centering
\begin{small}
\begin{tabular}{|c|c|c|c|c|c|c|}
\toprule
\begin{minipage}{0.1\linewidth}Model size (B)\end{minipage}& Data Selection & \begin{minipage}{0.1\linewidth}Translation Mode \end{minipage} & En $\rightarrow$ High & High $\rightarrow$ En & En $\rightarrow$ Low & Low $\rightarrow$ En \\
\midrule
1.2 & LM (100\%) & O & 8.96 & 15.76 & 0.65 & 3.01 \\
\midrule
1.2 & LM (90\%) -- MT (10\%) & C & 6.66 &	9.62 &	2.54 &	4.12 \\ 
1.2 & LM (90\%) -- MT (10\%) & O & 12.00 & 20.80 & 2.00 & 5.99 \\ 
\midrule
1.2 & LM (50\%) -- MT (50\%) & C & 14.68 &14.98 &	15.09 &	10.31 \\
1.2 & LM (50\%) -- MT (50\%) & O & 17.74 & 27.14 & 5.22 & 13.94 \\
\midrule
1.2 & EXP3 & C & 15.23 &	15.76 &	13.00 &	9.52 \\
1.2 & EXP3 & O & 16.05 & 26.20 & 4.63 & 18.07 \\
\midrule
1.2 & FAIR & C & 22.44 &	19.81 &	34.18 &	16.84 \\
1.2 & FAIR & O & 23.19 & 31.81 & 15.38 & 26.73 \\
\midrule
3.8 & LM (100\%) & O & 12.43 & 21.02 & 1.08 & 5.08 \\
\midrule
3.8 & EXP3 & C & 19.89 &	21.50 &	34.01 &	18.63 \\
3.8 & EXP3 & O & 26.63 & 34.88 & 23.63 & 31.53 \\
\midrule
3.8 & FAIR & C & 21.13 &	15.95 &	32.82 &	16.20 \\
3.8 & FAIR & O & 30.48 & 36.63 & 27.53 & 36.05 \\
\bottomrule
\end{tabular}
\end{small}
\caption{Comparison of Translation with control tokens (C) vs the one-shot setup (O) on MT tasks (Flores) measured with sacreBLEU. The one-shot setup yields better results except in the En $\rightarrow$ Low setting in which using control tokens is better.}
\label{tab:mt_super_vs_oneshot}
\end{table*}

\textbf{Translating with control tokens.} Recall that parallel data was used in a supervised fashion with the MT objective. For each language, a special control token was prefixed to the source sentence~\eqref{eq:mt_ingestion}. Such language control tokens
do not appear in the LM training data; therefore a natural question is whether supplying data to the pre-trained model in the form~\eqref{eq:mt_ingestion} results in translations to the desired language. This is indeed the case: \emph{at inference time the pre-trained LM performs the supervised task corresponding to each language control token}.

\textbf{Better translations are generated with in-context learning than by using control tokens.} A natural question is whether the resulting models produce better translations with in-context learning or by using control tokens. In Table~\ref{tab:mt_super_vs_oneshot} we compare the ``Translation Mode'' with control tokens (C) to the one with in-context learning. For in-context learning we use the one-shot setup (O), see the Appendix for examples of how the task is formulated. We clearly see that \emph{the one-shot setup outperforms the one with control tokens, except in the En $\rightarrow$ Low setting, in which the second is to be preferred}.

\textbf{Comparison to MTM and LLM baselines.} We compare our results to those reported in~\cite{xglm}, using their same spBLEU implementation to make the comparison fair. The first baseline, M2M~\cite{m2m} is an MTM trained with cross-lingual supervision.
The other two baselines, XGLM~\cite{xglm} and GPT-3~\cite{gpt3} consist of LLMs which  are trained on a self-supervised objective without any extra crosslingual supervision, and translations are obtained using in-context learning in the few-shot setup.

We do a comparison (Table~\ref{tab:ext_comparison_included}) considering those language pairs for which we also had parallel data. \emph{A clear advantage of our models is that we can always take the best score between translations generated with control tokens and in-context learning (one-shot): except for En$\rightarrow$Fi,
we outperform all the other models both at 1.2B and 3.8B parameters scale}.

\begin{table*}
\centering
\begin{small}
\begin{tabular}{|l|c|c|c|c|c|c|c|c|}
\toprule
Model            &  \begin{minipage}{0.1\linewidth} Translation Mode\end{minipage}  &   Ar$\rightarrow$En & De$\rightarrow$En & En$\rightarrow$Ar & En$\rightarrow$De & En$\rightarrow$Fi & En$\rightarrow$Fr & En$\rightarrow$Ko \\
\midrule																															
M2M-124 (0.6B)   &  -         &  25.5    & 35.8   & 17.9   & 32.6   & \textbf{24.2}  &  42.0  &  18.5       \\
GPT-3 (6.7B)     &  -          &  10.5    & 40.4   & 1.1    & 25.9   & 10.2   &  36.1  &  1.2    \\
XGLM (7.5B)      &  -          &  27.7    & 38.8   & 11.5   & 27.6   & 23.3   &  36.0  &  12.0       \\
Ours (1.2B)      &  C         &  20.2    & 32.9   & 18.6   & 34.4   & 18.7   &  45.4  &  10.7    \\
Ours (1.2B)      &  O          &  35.0    & 44.2   & 19.1   & 28.7   & 13.2   &  40.1  &  14.5    \\
Ours (3.8B)      &  C         &  15.1    & 24.6   & 20.5   & 27.4   & 19.4   &  34.7  &  10.8        \\
Ours (3.8B)      &  O          &  \textbf{41.2}    & \textbf{46.9}  & \textbf{25.5}   & \textbf{39.0}   & 21.2  &  \textbf{49.9}  &  \textbf{22.9}   \\
\midrule
Model            & \begin{minipage}{0.1\linewidth} Translation Mode\end{minipage}  &   En$\rightarrow$Ru & En$\rightarrow$Sw & Fi$\rightarrow$En & Fr$\rightarrow$En & Ko$\rightarrow$En & Ru$\rightarrow$En & Sw$\rightarrow$En \\
\midrule
M2M-124 (0.6B) &  - & 27.1 &	26.9 &	27.2 &	37.2 &	20.9 &	27.5 &	30.4 \\
GPT-3 (6.7B)     &  - & 11.2 &	0.5	& 25.3 &	42.8 &	8.3	& 28.1 &	5.0 \\
XGLM (7.5B) & - & 24.2 &	18.0 &	29.2 &	40.4 &	19.9 &	30.4 &	31.6 \\
Ours (1.2B) & C & 23.4 &	29.8 &	16.9 &	35.5 &	10.7 &	22.8 &	16.2 \\
Ours (1.2B) & O & 24.1 &	16.7 &	27.8 &	45.5 &	27.6 &	35.0 &	28.9 \\
Ours (3.8B) & C & 20.6 &	\textbf{34.7} &	18.6 &	25.6 &	10.2 &	20.1 &	15.6 \\
Ours (3.8B) & O & \textbf{31.6} &	26.7 &	\textbf{35.5} &	\textbf{47.6} &	\textbf{33.1} &	\textbf{37.5} &	\textbf{40.3} \\
\bottomrule
\end{tabular}
\end{small}
\caption{Comparison to MT baselines, for language pairs for which we included parallel data during training (Flores / spmBLEU).
For each language pair, selection of the best strategy between control tokens (C) and in-context learning (O) allows to
outperform the baselines.}
\label{tab:ext_comparison_included}
\end{table*}

We also looked at those language pairs reported in~\cite{xglm} for which we did not have parallel data. For the languages My and Ta our LM dataset had very little data and our models are unable to translate. For Ca, Bg and Hi we found that our systems do generate translations; \emph{when the target language is English, performance can be quite good as we outperform
the other systems on the directions Bg$\rightarrow$En, Hi$\rightarrow$En and Zh$\rightarrow$En}. Details of this evaluation are reported in the Appendix.

\begin{table*}
\centering
\begin{small}
\begin{tabular}{|l|c|c|c|c|c|c|c|c|}
\toprule
Data Sel. & Parallel as & TyDi QA & XTyDi QA & En $\rightarrow$ High & High $\rightarrow$ En & En $\rightarrow$ Low & Low $\rightarrow$ En & Wikilingua \\
\midrule
LM & -- & 25.59 & 10.40 & 8.96 & 15.76 & 0.65 & 3.10 & 12.71 \\ 
\midrule
EXP3 & MT & 31.39 & 16.54 & 16.05 & 26.20 & 4.63 & 18.07 & 12.57 \\
EXP3 & LM & 29.58 & 12.67 & 10.34 & 16.81 & 1.22 & 4.07 & 11.93 \\
\midrule
FAIR & MT & 32.19 & 18.85 & 23.19 & 31.81 & 15.38 & 26.73 & 11.82 \\
FAIR & LM & 18.06 & 8.53 & 8.21 & 15.15 & 1.89 & 5.73 & 10.35 \\
\bottomrule
\end{tabular}
\end{small}
\caption{Ablation (at model size 1.2B) of using the parallel data with the LM objective. For each data selection strategy,
using the non-English size of the parallel data with the LM objective significantly reduces performance on (X)TyDi QA and Translation.}
\label{tab:mt_as_lm_ablation}
\end{table*}

\subsection{Are the gains due to using more multilingual data?} Given that about $77\%$ of the data from \cite{palm} is in English, \emph{a natural conjecture is that adding parallel data is beneficial because it increases the non-English fraction of the data}. To test this hypothesis we construct  a new data-set by taking the non-English side of the MT data and applying to it the LM objective. We still use automated curriculum learning
to balance the two fractions of the LM data. In Table~\ref{tab:mt_as_lm_ablation} we compare the two approaches and \emph{observe a significant performance drop on both Question Answering and Machine Translation when using the MT data with the LM objective}. We conjecture that our MT data might be less useful at modeling language compared to the more rich kind of data from \cite{palm}.

\section{Conclusions}
We have demonstrated that, when \emph{pre-training Encoder-Decoder large language models}, the inclusion of cross-lingual supervision in the training objective is beneficial. In particular, we have found substantial gains when evaluating the resulting models on Machine Translation and Question Answering. One drawback of including parallel data is the introduction of a new hyper-parameter which quantifies the percentange of such data to use. Even though \emph{inclusion of some cross-lingual supervision is beneficial, determining the optimal amount by a grid search is unfeasible}; however, we have demonstrated that one can get good results by employing automated curriculum learning with multi-armed bandits~\cite{graves_curriculum}. Moreover, in our proposed approach the learned percentage can adjust during training and outperform the static data sampling baselines of~\cite{kale-etal-2021-nmt5}.

\section{Limitations}
Because of computing limitations, we investigated only Encoder-Decoder models. Further experiments are needed to extend the findings to Decoder-only models. Our summarization evaluations indicate improvements with increasing the parameter count, so further experiments with larger models (say $>8B$ parameters) might be needed to quantify the gains of adding parallel data during pre-training more precisely. Finally, while automated curriculum learning outperformed simple static data sampling strategies, more sophisticated sampling approaches might yield better results.

\bibliography{anthology,custom,example_paper.bib}

\begin{thebibliography}{20}
\expandafter\ifx\csname natexlab\endcsname\relax\def\natexlab#1{#1}\fi

\bibitem[{Auer et~al.(2002)Auer, Cesa-Bianchi, Freund, and
  Schapire}]{cesa_bianchi}
Peter Auer, Nicol\`{o} Cesa-Bianchi, Yoav Freund, and Robert~E. Schapire. 2002.
\newblock \href {https://doi.org/10.1137/S0097539701398375} {The nonstochastic
  multiarmed bandit problem}.
\newblock \emph{SIAM Journal on Computing}, 32(1):48--77.

\bibitem[{Brown et~al.(2020)Brown, Mann, Ryder, Subbiah, Kaplan, Dhariwal,
  Neelakantan, Shyam, Sastry, Askell, Agarwal, Herbert-Voss, Krueger, Henighan,
  Child, Ramesh, Ziegler, Wu, Winter, Hesse, Chen, Sigler, Litwin, Gray, Chess,
  Clark, Berner, McCandlish, Radford, Sutskever, and Amodei}]{gpt3}
Tom Brown, Benjamin Mann, Nick Ryder, Melanie Subbiah, Jared~D Kaplan, Prafulla
  Dhariwal, Arvind Neelakantan, Pranav Shyam, Girish Sastry, Amanda Askell,
  Sandhini Agarwal, Ariel Herbert-Voss, Gretchen Krueger, Tom Henighan, Rewon
  Child, Aditya Ramesh, Daniel Ziegler, Jeffrey Wu, Clemens Winter, Chris
  Hesse, Mark Chen, Eric Sigler, Mateusz Litwin, Scott Gray, Benjamin Chess,
  Jack Clark, Christopher Berner, Sam McCandlish, Alec Radford, Ilya Sutskever,
  and Dario Amodei. 2020.
\newblock \href
  {https://proceedings.neurips.cc/paper/2020/file/1457c0d6bfcb4967418bfb8ac142f64a-Paper.pdf}
  {Language models are few-shot learners}.
\newblock In \emph{Advances in Neural Information Processing Systems},
  volume~33, pages 1877--1901. Curran Associates, Inc.

\bibitem[{Chi et~al.(2021)Chi, Dong, Ma, Huang, Singhal, Mao, Huang, Song, and
  Wei}]{chi-etal-2021-mt6}
Zewen Chi, Li~Dong, Shuming Ma, Shaohan Huang, Saksham Singhal, Xian-Ling Mao,
  Heyan Huang, Xia Song, and Furu Wei. 2021.
\newblock \href {https://doi.org/10.18653/v1/2021.emnlp-main.125} {m{T}6:
  Multilingual pretrained text-to-text transformer with translation pairs}.
\newblock In \emph{Proceedings of the 2021 Conference on Empirical Methods in
  Natural Language Processing}, pages 1671--1683, Online and Punta Cana,
  Dominican Republic. Association for Computational Linguistics.

\bibitem[{Chowdhery et~al.(2022)Chowdhery, Narang, Devlin, Bosma, Mishra,
  Roberts, Barham, Chung, Sutton, Gehrmann, Schuh, Shi, Tsvyashchenko, Maynez,
  Rao, Barnes, Tay, Shazeer, Prabhakaran, Reif, Du, Hutchinson, Pope, Bradbury,
  Austin, Isard, Gur-Ari, Yin, Duke, Levskaya, Ghemawat, Dev, Michalewski,
  Garcia, Misra, Robinson, Fedus, Zhou, Ippolito, Luan, Lim, Zoph, Spiridonov,
  Sepassi, Dohan, Agrawal, Omernick, Dai, Pillai, Pellat, Lewkowycz, Moreira,
  Child, Polozov, Lee, Zhou, Wang, Saeta, Diaz, Firat, Catasta, Wei,
  Meier-Hellstern, Eck, Dean, Petrov, and Fiedel}]{palm}
Aakanksha Chowdhery, Sharan Narang, Jacob Devlin, Maarten Bosma, Gaurav Mishra,
  Adam Roberts, Paul Barham, Hyung~Won Chung, Charles Sutton, Sebastian
  Gehrmann, Parker Schuh, Kensen Shi, Sasha Tsvyashchenko, Joshua Maynez,
  Abhishek Rao, Parker Barnes, Yi~Tay, Noam Shazeer, Vinodkumar Prabhakaran,
  Emily Reif, Nan Du, Ben Hutchinson, Reiner Pope, James Bradbury, Jacob
  Austin, Michael Isard, Guy Gur-Ari, Pengcheng Yin, Toju Duke, Anselm
  Levskaya, Sanjay Ghemawat, Sunipa Dev, Henryk Michalewski, Xavier Garcia,
  Vedant Misra, Kevin Robinson, Liam Fedus, Denny Zhou, Daphne Ippolito, David
  Luan, Hyeontaek Lim, Barret Zoph, Alexander Spiridonov, Ryan Sepassi, David
  Dohan, Shivani Agrawal, Mark Omernick, Andrew~M. Dai,
  Thanumalayan~Sankaranarayana Pillai, Marie Pellat, Aitor Lewkowycz, Erica
  Moreira, Rewon Child, Oleksandr Polozov, Katherine Lee, Zongwei Zhou, Xuezhi
  Wang, Brennan Saeta, Mark Diaz, Orhan Firat, Michele Catasta, Jason Wei,
  Kathy Meier-Hellstern, Douglas Eck, Jeff Dean, Slav Petrov, and Noah Fiedel.
  2022.
\newblock \href {https://doi.org/10.48550/ARXIV.2204.02311} {Palm: Scaling
  language modeling with pathways}.

\bibitem[{Clark et~al.(2020)Clark, Choi, Collins, Garrette, Kwiatkowski,
  Nikolaev, and Palomaki}]{clark-etal-2020-tydi}
Jonathan~H. Clark, Eunsol Choi, Michael Collins, Dan Garrette, Tom Kwiatkowski,
  Vitaly Nikolaev, and Jennimaria Palomaki. 2020.
\newblock \href {https://doi.org/10.1162/tacl_a_00317} {{T}y{D}i {QA}: A
  benchmark for information-seeking question answering in typologically diverse
  languages}.
\newblock \emph{Transactions of the Association for Computational Linguistics},
  8:454--470.

\bibitem[{Fan et~al.(2021)Fan, Bhosale, Schwenk, Ma, El-Kishky, Goyal, Baines,
  Celebi, Wenzek, Chaudhary, Goyal, Birch, Liptchinsky, Edunov, Auli, and
  Joulin}]{m2m}
Angela Fan, Shruti Bhosale, Holger Schwenk, Zhiyi Ma, Ahmed El-Kishky,
  Siddharth Goyal, Mandeep Baines, Onur Celebi, Guillaume Wenzek, Vishrav
  Chaudhary, Naman Goyal, Tom Birch, Vitaliy Liptchinsky, Sergey Edunov,
  Michael Auli, and Armand Joulin. 2021.
\newblock \href {http://jmlr.org/papers/v22/20-1307.html} {Beyond
  english-centric multilingual machine translation}.
\newblock \emph{Journal of Machine Learning Research}, 22(107):1--48.

\bibitem[{Gehrmann et~al.(2022)Gehrmann, Bhattacharjee, Mahendiran, Wang,
  Papangelis, Madaan, McMillan-Major, Shvets, Upadhyay, Yao, Wilie,
  Bhagavatula, You, Thomson, Garbacea, Wang, Deutsch, Xiong, Jin, Gkatzia,
  Radev, Clark, Durmus, Ladhak, Ginter, Winata, Strobelt, Hayashi, Novikova,
  Kanerva, Chim, Zhou, Clive, Maynez, Sedoc, Juraska, Dhole, Chandu,
  Perez-Beltrachini, Ribeiro, Tunstall, Zhang, Pushkarna, Creutz, White, Kale,
  Eddine, Daheim, Subramani, Dusek, Liang, Ammanamanchi, Zhu, Puduppully, Kriz,
  Shahriyar, Cardenas, Mahamood, Osei, Cahyawijaya, Stajner, Montella, Jolly,
  Mille, Hasan, Shen, AMahidewumi, Raunak, Raheja, Nikolaev, Tsai, Jernite, Xu,
  Sang, Liu, and Hou}]{gem_benchmark}
Sebastian Gehrmann, Abhik Bhattacharjee, Abinaya Mahendiran, Alex Wang,
  Alexandros Papangelis, Aman Madaan, Angelina McMillan-Major, Anna Shvets,
  Ashish Upadhyay, Bingsheng Yao, Bryan Wilie, Chandra Bhagavatula, Chaobin
  You, Craig Thomson, Cristina Garbacea, Dakuo Wang, Daniel Deutsch, Deyi
  Xiong, Di~Jin, Dimitra Gkatzia, Dragomir Radev, Elizabeth Clark, Esin Durmus,
  Faisal Ladhak, Filip Ginter, Genta~Indra Winata, Hendrik Strobelt, Hiroaki
  Hayashi, Jekaterina Novikova, Jenna Kanerva, Jenny Chim, Jiawei Zhou, Jordan
  Clive, Joshua Maynez, João Sedoc, Juraj Juraska, Kaustubh~D. Dhole,
  Khyathi~Raghavi Chandu, Laura Perez-Beltrachini, Leonardo F.~R. Ribeiro,
  Lewis Tunstall, Li~Zhang, Mahima Pushkarna, Mathias Creutz, Michael White,
  Mihir~Sanjay Kale, Moussa~Kamal Eddine, Nico Daheim, Nishant Subramani,
  Ondrej Dusek, Paul~Pu Liang, Pawan~Sasanka Ammanamanchi, Qi~Zhu, Ratish
  Puduppully, Reno Kriz, Rifat Shahriyar, Ronald Cardenas, Saad Mahamood,
  Salomey Osei, Samuel Cahyawijaya, Sanja Stajner, Sébastien Montella, Shailza
  Jolly, Simon Mille, Tahmid Hasan, Tianhao Shen, Tosin~P. AMahidewumi, Vikas
  Raunak, Vipul Raheja, Vitaly Nikolaev, Vivian Tsai, Yacine Jernite, Ying Xu,
  Yisi Sang, Yixin Liu, and Yufang Hou. 2022.
\newblock \href {https://doi.org/10.48550/arXiv.2206.11249} {Gemv2:
  Multilingual nlg benchmarking in a single line of code}.
\newblock \emph{CoRR}, abs/2206.11249.

\bibitem[{Graves et~al.(2017)Graves, Bellemare, Menick, Munos, and
  Kavukcuoglu}]{graves_curriculum}
Alex Graves, Marc~G. Bellemare, Jacob Menick, R{\'e}mi Munos, and Koray
  Kavukcuoglu. 2017.
\newblock \href {https://proceedings.mlr.press/v70/graves17a.html} {Automated
  curriculum learning for neural networks}.
\newblock In \emph{Proceedings of the 34th International Conference on Machine
  Learning}, volume~70 of \emph{Proceedings of Machine Learning Research},
  pages 1311--1320. PMLR.

\bibitem[{Guzm{\'a}n et~al.(2019)Guzm{\'a}n, Chen, Ott, Pino, Lample, Koehn,
  Chaudhary, and Ranzato}]{guzman-etal-2019-flores}
Francisco Guzm{\'a}n, Peng-Jen Chen, Myle Ott, Juan Pino, Guillaume Lample,
  Philipp Koehn, Vishrav Chaudhary, and Marc{'}Aurelio Ranzato. 2019.
\newblock \href {https://doi.org/10.18653/v1/D19-1632} {The {FLORES} evaluation
  datasets for low-resource machine translation: {N}epali{--}{E}nglish and
  {S}inhala{--}{E}nglish}.
\newblock In \emph{Proceedings of the 2019 Conference on Empirical Methods in
  Natural Language Processing and the 9th International Joint Conference on
  Natural Language Processing (EMNLP-IJCNLP)}, pages 6098--6111, Hong Kong,
  China. Association for Computational Linguistics.

\bibitem[{Kale et~al.(2021)Kale, Siddhant, Al-Rfou, Xue, Constant, and
  Johnson}]{kale-etal-2021-nmt5}
Mihir Kale, Aditya Siddhant, Rami Al-Rfou, Linting Xue, Noah Constant, and
  Melvin Johnson. 2021.
\newblock \href {https://doi.org/10.18653/v1/2021.acl-short.87} {nm{T}5 - is
  parallel data still relevant for pre-training massively multilingual language
  models?}
\newblock In \emph{Proceedings of the 59th Annual Meeting of the Association
  for Computational Linguistics and the 11th International Joint Conference on
  Natural Language Processing (Volume 2: Short Papers)}, pages 683--691,
  Online. Association for Computational Linguistics.

\bibitem[{Kreutzer et~al.(2021)Kreutzer, Vilar, and
  Sokolov}]{kreutzer-etal-2021-bandits-dont}
Julia Kreutzer, David Vilar, and Artem Sokolov. 2021.
\newblock \href {https://doi.org/10.18653/v1/2021.findings-emnlp.274} {Bandits
  don{'}t follow rules: Balancing multi-facet machine translation with
  multi-armed bandits}.
\newblock In \emph{Findings of the Association for Computational Linguistics:
  EMNLP 2021}, pages 3190--3204, Punta Cana, Dominican Republic. Association
  for Computational Linguistics.

\bibitem[{Ladhak et~al.(2020)Ladhak, Durmus, Cardie, and
  McKeown}]{ladhak-etal-2020-wikilingua}
Faisal Ladhak, Esin Durmus, Claire Cardie, and Kathleen McKeown. 2020.
\newblock \href {https://doi.org/10.18653/v1/2020.findings-emnlp.360}
  {{W}iki{L}ingua: A new benchmark dataset for cross-lingual abstractive
  summarization}.
\newblock In \emph{Findings of the Association for Computational Linguistics:
  EMNLP 2020}, pages 4034--4048, Online. Association for Computational
  Linguistics.

\bibitem[{Lattimore and Szepesvári(2020)}]{lattimore_szepesvaari_2020}
Tor Lattimore and Csaba Szepesvári. 2020.
\newblock \href {https://doi.org/10.1017/9781108571401} {\emph{Bandit
  Algorithms}}.
\newblock Cambridge University Press.

\bibitem[{{Lin} et~al.(2021){Lin}, {Mihaylov}, {Artetxe}, {Wang}, {Chen},
  {Simig}, {Ott}, {Goyal}, {Bhosale}, {Du}, {Pasunuru}, {Shleifer}, {Singh
  Koura}, {Chaudhary}, {O'Horo}, {Wang}, {Zettlemoyer}, {Kozareva}, {Diab},
  {Stoyanov}, and {Li}}]{xglm}
Xi~Victoria {Lin}, Todor {Mihaylov}, Mikel {Artetxe}, Tianlu {Wang}, Shuohui
  {Chen}, Daniel {Simig}, Myle {Ott}, Naman {Goyal}, Shruti {Bhosale}, Jingfei
  {Du}, Ramakanth {Pasunuru}, Sam {Shleifer}, Punit {Singh Koura}, Vishrav
  {Chaudhary}, Brian {O'Horo}, Jeff {Wang}, Luke {Zettlemoyer}, Zornitsa
  {Kozareva}, Mona {Diab}, Veselin {Stoyanov}, and Xian {Li}. 2021.
\newblock \href {http://arxiv.org/abs/2112.10668} {{Few-shot Learning with
  Multilingual Language Models}}.
\newblock \emph{arXiv e-prints (to appear in EMNLP)}, page arXiv:2112.10668.

\bibitem[{Raffel et~al.(2019)Raffel, Shazeer, Roberts, Lee, Narang, Matena,
  Zhou, Li, and Liu}]{t5paper}
Colin Raffel, Noam Shazeer, Adam Roberts, Katherine Lee, Sharan Narang, Michael
  Matena, Yanqi Zhou, Wei Li, and Peter~J. Liu. 2019.
\newblock \href {http://arxiv.org/abs/1910.10683} {Exploring the limits of
  transfer learning with a unified text-to-text transformer}.
\newblock \emph{arXiv e-prints}.

\bibitem[{Reid and Artetxe(2022)}]{reid-artetxe-2022-paradise}
Machel Reid and Mikel Artetxe. 2022.
\newblock \href {https://doi.org/10.18653/v1/2022.naacl-main.58} {{PARADISE}:
  Exploiting parallel data for multilingual sequence-to-sequence pretraining}.
\newblock In \emph{Proceedings of the 2022 Conference of the North American
  Chapter of the Association for Computational Linguistics: Human Language
  Technologies}, pages 800--810, Seattle, United States. Association for
  Computational Linguistics.

\bibitem[{Roberts et~al.(2022)Roberts, Chung, Levskaya, Mishra, Bradbury,
  Andor, Narang, Lester, Gaffney, Mohiuddin, Hawthorne, Lewkowycz, Salcianu,
  van Zee, Austin, Goodman, Soares, Hu, Tsvyashchenko, Chowdhery, Bastings,
  Bulian, Garcia, Ni, Chen, Kenealy, Clark, Lee, Garrette, Lee-Thorp, Raffel,
  Shazeer, Ritter, Bosma, Passos, Maitin-Shepard, Fiedel, Omernick, Saeta,
  Sepassi, Spiridonov, Newlan, and Gesmundo}]{roberts2022t5x}
Adam Roberts, Hyung~Won Chung, Anselm Levskaya, Gaurav Mishra, James Bradbury,
  Daniel Andor, Sharan Narang, Brian Lester, Colin Gaffney, Afroz Mohiuddin,
  Curtis Hawthorne, Aitor Lewkowycz, Alex Salcianu, Marc van Zee, Jacob Austin,
  Sebastian Goodman, Livio~Baldini Soares, Haitang Hu, Sasha Tsvyashchenko,
  Aakanksha Chowdhery, Jasmijn Bastings, Jannis Bulian, Xavier Garcia, Jianmo
  Ni, Andrew Chen, Kathleen Kenealy, Jonathan~H. Clark, Stephan Lee, Dan
  Garrette, James Lee-Thorp, Colin Raffel, Noam Shazeer, Marvin Ritter, Maarten
  Bosma, Alexandre Passos, Jeremy Maitin-Shepard, Noah Fiedel, Mark Omernick,
  Brennan Saeta, Ryan Sepassi, Alexander Spiridonov, Joshua Newlan, and Andrea
  Gesmundo. 2022.
\newblock \href {https://arxiv.org/abs/2203.17189} {Scaling up models and data
  with $\texttt{t5x}$ and $\texttt{seqio}$}.
\newblock \emph{arXiv preprint arXiv:2203.17189}.

\bibitem[{Tay et~al.(2022)Tay, Dehghani, Tran, Garcia, Wei, Wang, Chung, Bahri,
  Schuster, Zheng, Zhou, Houlsby, and
  Metzler}]{https://doi.org/10.48550/arxiv.2205.05131}
Yi~Tay, Mostafa Dehghani, Vinh~Q. Tran, Xavier Garcia, Jason Wei, Xuezhi Wang,
  Hyung~Won Chung, Dara Bahri, Tal Schuster, Huaixiu~Steven Zheng, Denny Zhou,
  Neil Houlsby, and Donald Metzler. 2022.
\newblock \href {https://doi.org/10.48550/ARXIV.2205.05131} {Ul2: Unifying
  language learning paradigms}.

\bibitem[{{Vilar} et~al.(2022){Vilar}, {Freitag}, {Cherry}, {Luo}, {Ratnakar},
  and {Foster}}]{palm_mt_prompt}
David {Vilar}, Markus {Freitag}, Colin {Cherry}, Jiaming {Luo}, Viresh
  {Ratnakar}, and George {Foster}. 2022.
\newblock \href {http://arxiv.org/abs/2211.09102} {{Prompting PaLM for
  Translation: Assessing Strategies and Performance}}.
\newblock \emph{arXiv e-prints}, page arXiv:2211.09102.

\bibitem[{Xue et~al.(2021)Xue, Constant, Roberts, Kale, Al-Rfou, Siddhant,
  Barua, and Raffel}]{mt5}
Linting Xue, Noah Constant, Adam Roberts, Mihir Kale, Rami Al-Rfou, Aditya
  Siddhant, Aditya Barua, and Colin Raffel. 2021.
\newblock \href {https://doi.org/10.18653/v1/2021.naacl-main.41} {m{T}5: A
  massively multilingual pre-trained text-to-text transformer}.
\newblock In \emph{Proceedings of the 2021 Conference of the North American
  Chapter of the Association for Computational Linguistics: Human Language
  Technologies}, pages 483--498, Online. Association for Computational
  Linguistics.

\end{thebibliography}
\bibliographystyle{acl_natbib}

\appendix

\section{Training Hyper-parameters}
We train the 1.2B models for 250k steps and the 3.8B models for 500k steps. We use T5X and SeqIO; the input sequences use packing with slightly over
500k non-padding tokens for each batch. The learning rate uses square-root decay, with a base learning rate of $1.0$ and 10k warm-up steps.
We use the default Adafactor optimizer of the T5X library.

\section{Technical details for using Automated Curriculum Learning}
Here we report a couple of crucial technical details for correctly using Curriculum Learning.

\cite{graves_curriculum, kreutzer-etal-2021-bandits-dont} use rescaled rewards:
\begin{enumerate}
    \item Keep a priority queue of the last $T$ rewards. While the queue is not $x\%$ full, return $0$ as a rescaled reward (so the bandit algorithm is not learning). 
    \item When the queue is $x\%$ full compute the  $20$-th and $80$-th quantiles.
    \item When a new reward $r$ comes is, clip it to lie between the current $20$-th and $80$-th quantiles to obtain $r'$; then linearly rescale $r'$
    to $r''\in[-1,1]$ where $-1$ corresponds to the  $20$-th  quantile and $+1$ to the $80$-th quantiles.
    \item Supply $r''$ to the  bandit algorithm.
    \item Enqueue $r$ and recompute the  $20$-th and $80$-th quantiles.
\end{enumerate}
Therefore, \cite{graves_curriculum, kreutzer-etal-2021-bandits-dont} use effective rewards in $[-1,1]$; however the proofs of convergence for EXP3 in \cite{cesa_bianchi, lattimore_szepesvaari_2020} do not work with negative rewards. Also FAIR can produce negative probability weights if rewards are negative.
We therefore rescale rewards so that $r''\in[0,1]$ and $0$ corresponds to the  $20$-th quantile of the queue. We use a queue of length $T=5000$ and $x=10\%$.

Note also that \cite{graves_curriculum} claims to use EXP3S; however, with their choice of parameters it always defaults to EXP3. We therefore do not mention EXP3S in this work as it might be confusing. 

\section{Hyper-parameters for Curriculum Learning}
For EXP3 we set the learning rate to $10^{-3}$ and the exploration rate to $25\%$.

For FAIR the exploration rate is set to $10\%$ and $\mu$ is set to $10^{-2}$. Note that $\mu$ operates on updating the 
moving average, so our choice corresponds to a time horizon of 100 steps.

\section{One-shot task format}
Our models are evaluated with in-context learning, using the one-shot paradigm. Here are examples of the formulation for the different tasks.

For Question Answering the input is of the form: ``\textbf{Context: The European jackal ...\textbackslash n\textbackslash n Q: How many jackals ... ?\textbackslash n A: 70,000\textbackslash n\textbackslash n} Context: The first known specimens of ... \textbackslash n\textbackslash n Q: When was the ... ? \textbackslash n A:''. The bold part is the single example supplied to use the in-context learning paradigm (here it's one-shot).

For Machine Translation the input is of the form: ``\textbf{German:  Am 28. Juni wurde Marshall Italo Balbo, ...\textbackslash n English: On June 28, Marshal Italo Balbo, ...\textbackslash n \textbackslash n} German:  Dr. Ehud Ur, Professor f\"ur Medizin ... \textbackslash n \textbackslash n English:''. The bold part is the single example supplied to use the in-context learning paradigm (here it's one-shot).

For Summarization the input is of the from: ``\textit{I will first show a set of step-by-step instructions and then write a short summary of every step in the same language of the instructions.} \textbackslash n \textbackslash n \textbf{Summarize the following instructions: Loneliness can take ... \textbackslash n Summary: Identify your type of loneliness. Realize that loneliness is a feeling. Consider your personality. Recognize that you are not alone in feeling lonely. \textbackslash n \textbackslash n} Summarize the following instructions: Usually, rainbows are ... \textbackslash n Summary:''. The bold part is the single example supplied to use the in-context learning paradigm (here it's one-shot). The part in italics is a prompt used by the GEM benchmark.

\section{Translating with Control Tokens}
Our models translate sentences to the desired languages when  the special language tokens are prefixed to the inputs. In this case the inputs have the following form: ``<2en> Dr. Ehud Ur, Professor f\"ur Medizin ...'', where ``<2en>'' denotes the task of translating to English.

\section{Comparison to MTM and LLM baselines for languages without parallel data}
We report the comparison to MTM and LLM baselines for those languages for which we did not have parallel data during training in Table~\ref{tab:ext_comparison_not_included}.

Note that our systems cannot be evaluated with language control tokens when the target language was absent from the parallel data supplied at training time; therefore we evaluate in the one-shot setup. Our systems struggle with the languages My and Ta which were extremely unrepresented in the LM data. For other languages,  we have 
indication that the model is able to translate. Moreover, performance on translation to English can be quite good, e.g.~outperforming the other systems especially on the pairs Bg$\rightarrow$En, Hi$\rightarrow$En and Zh$\rightarrow$En.

Note also that the model trained with FAIR outperforms the model trained with LM data only. Therefore, parallel data has improved translations also to language pairs that were not present in the cross-lingual supervised data.

\begin{table*}
\centering
\begin{small}
\begin{tabular}{|l|c|c|c|c|c|c|}
\toprule
Model &   Bg$\rightarrow$En & Ca$\rightarrow$En & En$\rightarrow$Bg & En$\rightarrow$Ca & En$\rightarrow$Hi & En$\rightarrow$My \\
\midrule																															
M2M-124 (0.6B)   &  33.0 &	33.4 &	\textbf{37.4} &	31.2 &	\textbf{28.1} &	3.5  \\
GPT-3 (6.7B)     & 21.6	& 40.2 & 5.9 &	23.8 &	0.3 &	0.1  \\
XGLM (7.5B)      &  35.5 &	\textbf{41.1} &	33.1 &	\textbf{34.0} &	19.9 &	\textbf{11.0 }    \\
Ours  (FAIR, 3.8B)      &  \textbf{36.9} &	39.4 &	16.8 &	19.3 &	8.0 &	0.4    \\
Ours (100\% LM, 3.8B)      &  25.3	& 29.7 &	12.1 &	14.9 &	3.3 &	0.4 \\
\midrule
Model            & En$\rightarrow$Ta & En$\rightarrow$Zh & Hi$\rightarrow$En & My$\rightarrow$En & Ta$\rightarrow$En & Zh$\rightarrow$En \\
\midrule
M2M-124 (0.6B) &  3.4 &	\textbf{19.3} &	27.9 &	10.0 &	8.3	& 20.9 \\
GPT-3 (6.7B)     &  0.0 &	12.5 &	1.2 &	0.5	 & 1 &	21.1 \\
XGLM (7.5B) &  \textbf{8.5} &	15.6 &	25.2 & \textbf{14.1} &	\textbf{16.3} &	20.7 \\
Ours (FAIR, 3.8B) &  0.4 &	12.0 &	\textbf{28.6} &	2.0 &	6.4 &	\textbf{25.4} \\
Ours LM only (100\% LM, 3.8B) & 0.4 &	8.0 &	11.6 &	2.0 &	3.4 &	17.1 \\
\bottomrule
\end{tabular}
\end{small}
\caption{Comparison to MT baselines, for language pairs for which we did not include parallel data during training (Flores / spmBLEU). 
In our stystems, adding parallel data improves one-shot translation performance also on language pairs that were not included in 
parallel data.}
\label{tab:ext_comparison_not_included}
\end{table*}

\section{Visualizing the rewards during training}
\begin{figure}
\centering
\centerline{\includegraphics[width=\columnwidth]{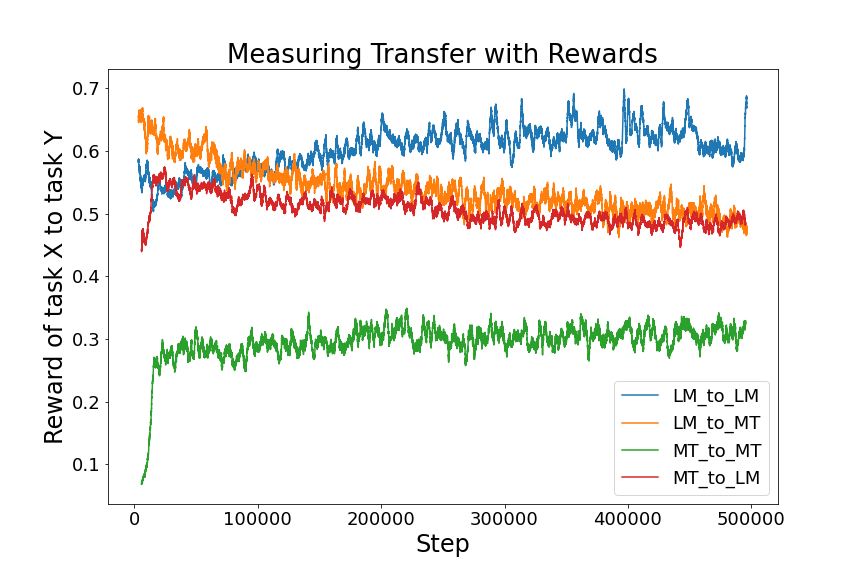}}
\caption{Measuring transfer as the reward on task Y resulting from a gradient step on task X (X to Y in the legend).
At the beginning of training the rewards of MT to itself is lower and increases over time. There is always a substantial transfer
from MT to LM. We use a running average with window 500 to reduce variance.
}
\label{fig:transfer}
\end{figure}

We plot in Figure~\ref{fig:transfer} the transfer between tasks, measured in terms of rewards. Specifically,
when we took a gradient step on task $X$ and evaluated the reward on task $Y$ we get a measure of transfer from $X$ to $Y$
that can be plotted over time. Note that there is always positive transfer from the MT to the LM objective; however the LM objective
has on average higher rewards when applied to MT or LM. Interestingly,
the transfer from MT to itself was low at the beginning of training and increased over time.

\end{document}